\documentclass{article}

\usepackage[preprint]{neurips_2019}

\usepackage[utf8]{inputenc} 
\usepackage[T1]{fontenc}    
\usepackage{hyperref}       
\usepackage{url}            
\usepackage{booktabs}       
\usepackage{amsfonts}       
\usepackage{nicefrac}       
\usepackage{microtype}      

\usepackage{graphicx}
\usepackage{caption}
\usepackage{subcaption}
\usepackage{booktabs} 
\usepackage{sidecap}
\usepackage{wrapfig}

\usepackage{amsmath}
\usepackage{amsthm}
\usepackage{dsfont}
\usepackage{amssymb}
\usepackage{amscd}
\usepackage{mathtools}
\mathtoolsset{showonlyrefs=true}

\usepackage{makecell}

\usepackage{hyperref}

\usepackage{xcolor}

\newcommand{\PAR}[1]{\vspace{-2pt} \noindent {\bf #1~}}

\title{\textsc{MuleX}: Disentangling\\Exploitation from Exploration in Deep RL}

\author{%
  \makecell{
  Lucas Beyer$^{\mathbf{1}}$,
  \, Damien Vincent$^{\mathbf{1}}$,
  \, Olivier Teboul$^{\mathbf{2}}$
  \\
  Sylvain Gelly$^{\mathbf{1}}$,
  \, Matthieu Geist$^{\mathbf{2}}$,
  \, Olivier Pietquin$^{\mathbf{2}}$}
  \\
  Google Research, Brain Team ($^{1}$Z\"urich $^{2}$Paris) \\
  \texttt{\{lbeyer, damienv, oliviert, sylvaingelly, mfgeist, pietquin\}@google.com} \\
}

\begin{document}

\maketitle

\begin{abstract}
An agent learning through interactions should balance its action selection process between probing the environment to discover new rewards and using the information acquired in the past to adopt useful behaviour.
This trade-off is usually obtained by perturbing either the agent's actions (\textit{e.g.}, $\epsilon$-greedy or Gibbs sampling) or the agent's parameters (\textit{e.g.}, NoisyNet), or by modifying the reward it receives (\textit{e.g.}, exploration bonus, intrinsic motivation, or hand-shaped rewards).
Here, we adopt a disruptive but simple and generic perspective, where we explicitly disentangle exploration and exploitation.
Different losses are optimized in parallel, one of them coming from the true objective (maximizing cumulative rewards from the environment) and others being related to exploration.
Every loss is used in turn to learn a policy that generates transitions, all shared in a single replay buffer.
Off-policy methods are then applied to these transitions to optimize each loss.
We showcase our approach on a hard-exploration environment, show its sample-efficiency and robustness, and discuss further implications.
\end{abstract}

\section{Introduction}
\label{sec:intro}

This paper addresses the ``exploration vs.\ exploitation'' dilemma arising in Reinforcement Learning (RL).
It suggests that an agent learning through interactions should balance its action selection process between probing the environment to discover new rewards (exploration) and using the information acquired in the past to adopt an acceptable behaviour (exploitation).
This trade-off is usually obtained by modifying the actions selected by the RL agent (\textit{e.g.}, $\epsilon$-greedy selection, Gibbs Sampling, optimism~\citep{NIPS2006_3052,geist2011managing}), by perturbing the parameters of the agent~\citep{fortunato2017noisy,plappert2017parameter}, or by modifying the reward it receives (\textit{e.g.}, exploration bonus or intrinsic motivation~\citep{bellemare2016unifying,tang2017exploration}).
Those methods often rely on many meta-parameters that are hard to tune, \textit{ad hoc} to the problem at hand and, most importantly, can lead to sub-optimal policies.

\begin{figure}
  \begin{center}
    \includegraphics[width=1.0\linewidth]{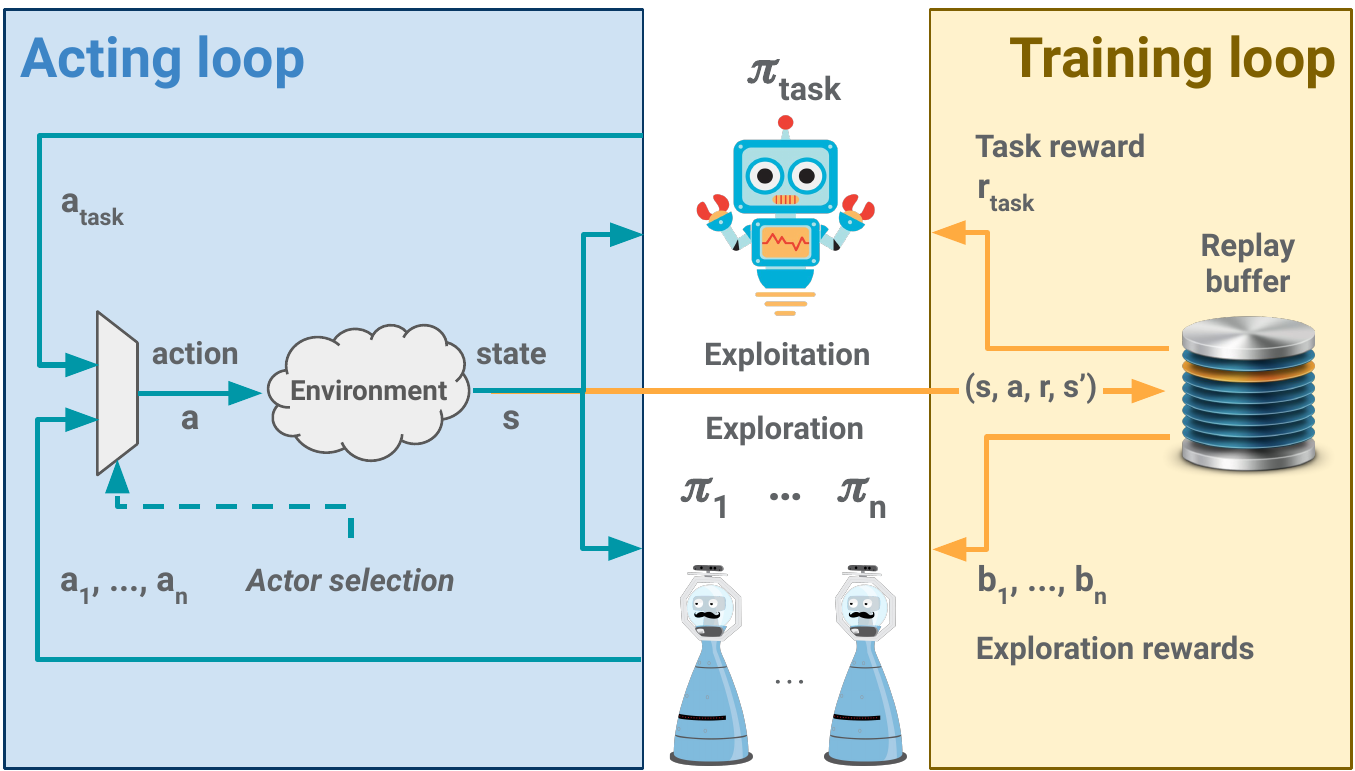}
  \end{center}
  \vspace*{-10pt}
  \caption{An overview of the proposed framework: a single agent plays with its environment while maintaining several policies.
  At a given step, it follows either the task policy or one of the exploration policies, disentangling exploration from exploitation.
  The task policy is trained with the rewards coming from the environment while exploration policies are learned through exploration rewards. 
  }\label{fig:xplor_overview}
  \vspace*{-10pt}
\end{figure}

Here, we adopt a disruptive but simple and generic perspective, where we disentangle explicitly exploration and exploitation in a deep RL architecture, as depicted in Figure~\ref{fig:xplor_overview}.
Different losses are optimized in parallel, one of them coming from the true RL objective (that is maximizing the cumulative rewards gathered in the environment) and others being related to exploration (\textit{e.g.}, exploration bonus or intrinsic motivation).
Every loss is used in turn to compute a policy that generates transitions, all shared in a single replay buffer.
Off-policy RL methods are then applied to these transitions to optimize every loss including the true RL one.
This approach is generic, as we can combine many existing exploration strategies, and make use of any off-policy RL algorithm.

After discussing related works, we present the general proposed strategy, which we call \textsc{MuleX} for ``Multiple losses for eXploration'', as well as a specific instantiation based on DQN with an agent combining an exploiter and an explorer optimizing for a count-based exploratory loss.
Then, we showcase this approach on a hard-exploration environment.
Notably, we show through an ablation study that the proposed approach is more efficient than any of its individual components, learns faster, and is more stable.

\section{Related work}
\label{sec:rel}

The exploration-exploitation dilemma is a core problem of RL.
Its simplest form is the stochastic bandit problem \citep{bubeck2012regret}, where an agent has to pull sequentially arms associated to stochastic rewards such as maximizing the expected cumulative reward.
In this case, the uncertainty comes only from the stochasticity of the rewards.
In RL, things are more involved, as a decision can have long-term consequences.

A common and simple approach is to disrupt the greedy action of the policy, for example with an $\epsilon$-greedy policy or with Gibbs sampling \citep{sutton2018reinforcement}.
These common approaches, if simple, are usually inefficient in hard exploration problems (for example, when rewards are scarce and far from the initial state).
An alternative approach consists in perturbing the parameters of the agent, instead of its actions \citep{sehnke2010parameter,plappert2017parameter,fortunato2017noisy}.
This comes with various motivations, such as performing a consistent exploration.
It has been recently shown that this kind of exploration has a lower sample complexity \citep{vemula2019contrasting}.
A third approach consists in enhancing the reward with an exploration bonus, that can be based on some form of intrinsic motivation, novelty measure, or expert knowledge.
For example, novelty can be measured by the number of times a state has been visited \citep{brafman2002r}, an approach that was successfully scaled to deep RL \citep{bellemare2016unifying,tang2017exploration}.
Other novelty measures such as prediction error \citep{burda2018exploration} have been proposed.
However, modifying the reward changes the problem at hand, and the bonus should generally be carefully annealed. 
Finally, optimism in the face of uncertainty, a well-known paradigm in online learning, has also been applied to RL~\citep{NIPS2006_3052,geist2011managing}.
It consists in a modification of the action-selection probability to increase the chance of selecting an action with upper confidence bound on the reward.
Yet, this is often computationally intractable as it requires computing second order statistics, either on the state visitation frequency, or the model parameters.

All aforementioned approaches share the property of having a single exploration strategy.
In some distributed RL algorithms, it is advocated that different workers use different parameters for their exploration strategy, such as the value of $\epsilon$ in an $\epsilon$-greedy strategy \citep{mnih2016asynchronous}.
Yet, the exploration strategy is homogeneous, and a single loss is optimized by the learner.
\citet{jaderberg2018human} propose a similar strategy in a population-based multi-agent approach.
Different agents have different (entangled) exploration strategies, and a second optimization process evolves them according to the true environment rewards.

What is common between all these methods is that they entangle exploration and exploitation into a single policy.
A recent exception was proposed by \citet{colas2018gep}.
They have first a pure exploration phase (based on intrinsic motivation) to feed a replay buffer that is then, in a second step, used to train the task policy with a classic off-policy algorithm (that has its own exploration mixed in).
The approach we propose is different, notably by the fact that we optimize for different losses at the same time: one for exploitation, and one or more for exploration. Our scheme results in distinct policies, which  interact for gathering samples. 

\section{Method}
\label{sec:method}

While \textsc{MuleX} is generally applicable to any type of off-policy RL agent, we describe an instantiation based on DQN \citep{mnih2015human} for simplicity and extend it to Rainbow in the Appendix. 

Consider the standard RL setting where an agent learns to solve a given Markov Decision Process (MDP) defined by the 5-tuple $\{S, A, P, r, \gamma\}$ of the state space, action space, transition function, reward function, and discount factor, respectively.
To solve this problem, the agent iterates between acting in the environment in order to collect $(s_i, a_i, r_i, s_{i+1})$ transitions which are stored in a replay buffer, and updating the function approximator for the $Q$-function using transitions from that replay buffer by optimizing:
\begin{equation}
    \hat{\mathbb{E}}\left[\big(Q(s_i,a_i) - Y(r_i, s_{i+1})\big)^2\right]\label{eq:dqn_update},
\end{equation}
where the target $Y$ is the bootstrap estimate of the optimal expected return:
\begin{equation}
    Y(r_i, s_{i+1}) = r_i + \gamma \max_{a'}Q_\text{target}(s_{i+1}, a').\label{eq:dqn_target}
\end{equation}
A policy $\pi_Q$ is associated to this $Q$-function by taking the best action according to $Q$ at each step.

The minimal way of doing exploration is by making the policy $\pi_Q$ $\epsilon$-greedy, meaning with probability $\epsilon$, a random action is taken instead of the optimal one.
While this theoretically finds the optimal policy in the limit, it does not work well on hard exploration tasks in practical settings where time is constrained.

A widespread approach to encourage more structured exploration is to augment the environment's reward $r_i$ with a bonus for exploration, be it through hand-engineered guiding rewards, intrinsic motivation, or curiosity.
These boni, denoted by $b^1_i, b^2_i, ...$, are typically added to the environment's reward with some weighting factors, resulting in a different target for the DQN optimization:%
\begin{equation}
    Y_\text{Add} \equiv Y(\alpha r_i + \beta^1 b^1_i + \beta^2 b^2_i + \cdots, s_{i+1}).\label{eq:dqn_additive_target}
\end{equation}
The weight $\beta$ needs to be tuned such that it does encourage exploration while simultaneously not drowning the actual task-based reward $r_i$.

Note that the $Q$-function learned using such modified rewards solves a  MDP being different from the original one, which in addition is usually non-stationary as the ``novelty'' of states changes over time.
This is an undesired side-effect of encouraging exploration through such boni.
A manually tuned annealing schedule is often imposed on $\beta$ for mitigating this, but even doing so, behaviours of exploration can still be observed in the final agent.

One especially prominent type of such bonus is derived from the bandits literature, where favourable theoretical bounds on regret have been shown~\citep{auer2002ucb} when adding $\sqrt{{2 \log i}/{N}}$ to an arm's estimated value, where $i$ is the total number of pulls, and $N$ is that arm's number of pulls.
This was translated into the RL setting \citep{Strehl08,bellemare2016unifying} by suggesting a bonus related to the number of times a configuration has been visited: $b_i = {1}/{\sqrt{N(s_i,a_i)}}$, with a visit counter $N(s,a)$.
The term $\sqrt{2 \log i}$ is almost constant and can be absorbed into the weight $\beta$.
One important difference between what is suggested in bandits and what is done in these papers, is that in the bandits literature, this bonus is \emph{not} added to an arm's estimated value, but it is only used for the action selection process.
One argument for adding the bonus to the reward is to encourage long term planning at visiting unexplored areas of the state space.
The bandit approach has been adopted in RL as well~\citep{NIPS2006_3052,geist2011managing} but this line of work does not immediately translate to the deep RL framework.

Our proposal is to disentangle these rewards in the agent by learning a separate $Q$-function for each of them:
\begin{align}
    Y_\text{task} &\equiv Y(r_i, s_{i+1}) \\
    Y_{b^1} &\equiv Y(b^1_i, s_{i+1}) \\
    Y_{b^2} &\equiv Y(b^2_i, s_{i+1}) \\
            &\vdotswithin{\equiv}
\end{align}
Note that all $Q$-functions are learned from the same, shared transitions.
This way, by acting according to each individual Q-function, we obtain one policy corresponding to each reward, which can be used to act according to that reward's intent.
Most importantly, $\pi_\text{task}$ attempts to solve the actual task through the whole training.

Multiple policies are thus available, some focused on exploration, one on solving the task, and the RL agent needs to decide with which one to act and collect transitions.
Many strategies can be conceived, including learning-based ones, but they should involve all policies (see Sec.~\ref{subsec:all_policies}).
In this work, we show that even using a simple random heuristic, our proposed framework has considerable benefits over typical methods relying on the sum of rewards to optimize the policy.
The heuristic works as follows.
First, choose which policy to use for acting according to a categorical distribution of parameters $(p_{\text{task}}, p_{b^1}, p_{b^2}, \dots)$.
Then, sample the number of steps for which this policy should act from a geometric distribution with parameter $(1 - \gamma_\text{steps})$.
After acting for that number of steps, rinse and repeat.
While setting $\gamma_\text{steps}$ to the MDP's $\gamma$ is a reasonable approach, we leave it as a free hyperparameter to be optimized.

\section{Experiments and results}
\label{sec:xp}

\begin{wrapfigure}{r}{0.5\textwidth}
    \begin{subfigure}[b]{0.49\linewidth}
        \begin{center}
            \includegraphics[width=0.9\linewidth]{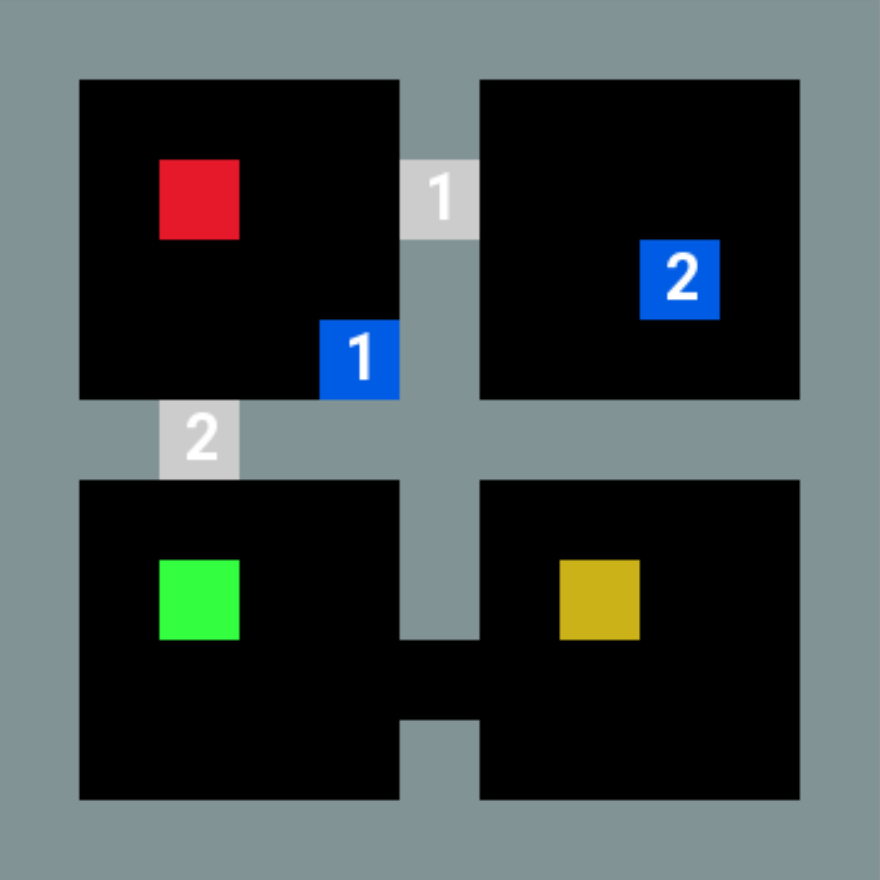}
            \caption{The full environment}
            \label{fig:montezuminha_full}
        \end{center}
    \end{subfigure}
    \begin{subfigure}[b]{0.49\linewidth}
        \begin{center}
            \includegraphics[width=0.9\linewidth]{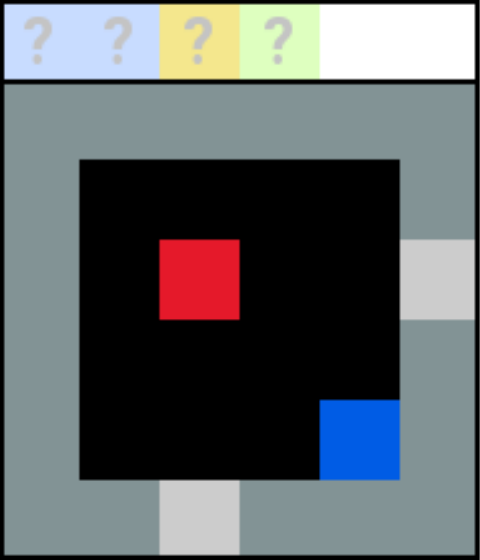}
            \caption{Agent's input}
            \label{fig:montezuminha_state}
        \end{center}
    \end{subfigure}
    \caption{The Montezuminha environment.}
    \label{fig:env}
\end{wrapfigure}

We experiment on a grid world environment inspired by two popular environments for exploration in RL research: Montezuma's Revenge and the classic four-rooms, but where we can explicitly control the various aspects which influence exploration.
The environment, which we call Montezuminha and bears similarity to the classic four-rooms, is shown in Figure~\ref{fig:env}.
The agent starts in the upper left one, where a key opens the door to the upper right one.
There, a second key opens another door in the first room leading to the third room (lower left).
There, the agent can either end the game by finding the exit or explore the last room (lower right) to get an extra reward.
Every collected item and reaching the exit give a +1 reward.
The maximum score is then +4, provided that the agent does avoid an early exit in the third room, which should require strong exploration or extreme luck.

With this, we can explicitly control for the various factors affecting exploration:
\begin{itemize}
    \item By increasing the size of the rooms, we increase sparsity of rewards. The number of steps along the optimal trajectory grows linearly, while the amount of exploration grows quadratically.
    \item By making the walls teleport the agent to a rewardless parallel world, we add some distracting states whose exploration is not aligned with solving the task.
    \item By adding deadly ghosts which move randomly, we add stochasticity to the environment. This is interesting since real tasks can be stochastic due to partial observability and nature.
    \item We can use an \emph{oracle} exploration bonus on the plain environment, or an approximate exploration bonus by rendering a textured version of the environment.
\end{itemize}

We perform our expriments in the Dopamine framework \citep{castro18dopamine}, where we train the agent during 800 iterations, each one consisting of 2500 training steps and then 1250 evaluation steps.
We limit each episode to 500 steps and perform gradient-updates using RMSProp on mini-batches of 32 transitions every 4 training steps.
Our neural network takes as input a stack of 4 consecutive frames (one frame is shown in Figure~\ref{fig:montezuminha_state}), and consists of a shared body with two convolutional layers with 16 and 32 kernels and then for each head two dense layers with 64 hidden neurons.
All reported scores are obtained by running the task-policy $\pi_\text{task}$ in the environment in ``evaluation mode,'' \textit{i.e.}\ without collecting transitions for training.

We focus on comparing our proposed \textsc{MuleX} approach to the typical \emph{Additive} approach which optimizes a single policy using as a reward a linear combination of the task and the bonus reward.
Our method is instantiated with two policies, one optimizing the task reward and one optimizing only the bonus reward.
Note that our purpose is not to compare how different exploration rewards perform, we thus assume that both methods have access to an \emph{oracle} exploration bonus based on exact state-counts $b_i = 1/\sqrt{N(s_i)}$.
Other works have proposed ways to extend such exploration bonuses to large problems \citep{bellemare2016unifying,tang2017exploration,burda2018exploration}.
As another baseline, we include experiments with $\epsilon${\it-greedy} as sole exploration method.

We perform a large-scale comparison between \textsc{MuleX}, {\it Additive}, and {\it $\epsilon\text{-greedy}$} agents using random hyperparameter search \citep{bergstra2012random}, providing the same budget of 200 trials (repeated 5 times each) to every method.
For the {\it $\epsilon\text{-greedy}$} agent, we logarithmically search over $\epsilon \in [0.001, 0.5]$.
For the {\it Additive} agent, we logarithmically search over the bonus weight $\beta \in [0.01, 100]$.
For the \textsc{MuleX} agent, we search over the switching strategy's start probabilities $p_\text{task} \in [0.5, 0.9]$ and duration $\gamma_\text{steps} \in [0.8, 0.99]$.
Note that there is no $\beta$ to be tuned in \textsc{MuleX}.
For all agents, we logarithmically search over the optimizer's learning-rate in $[1\mathrm{e}{-5}, 1\mathrm{e}{-3}]$.

\begin{figure}[t]
\begin{minipage}[t]{0.49\linewidth}
  \begin{center}
    \includegraphics[width=1.0\linewidth]{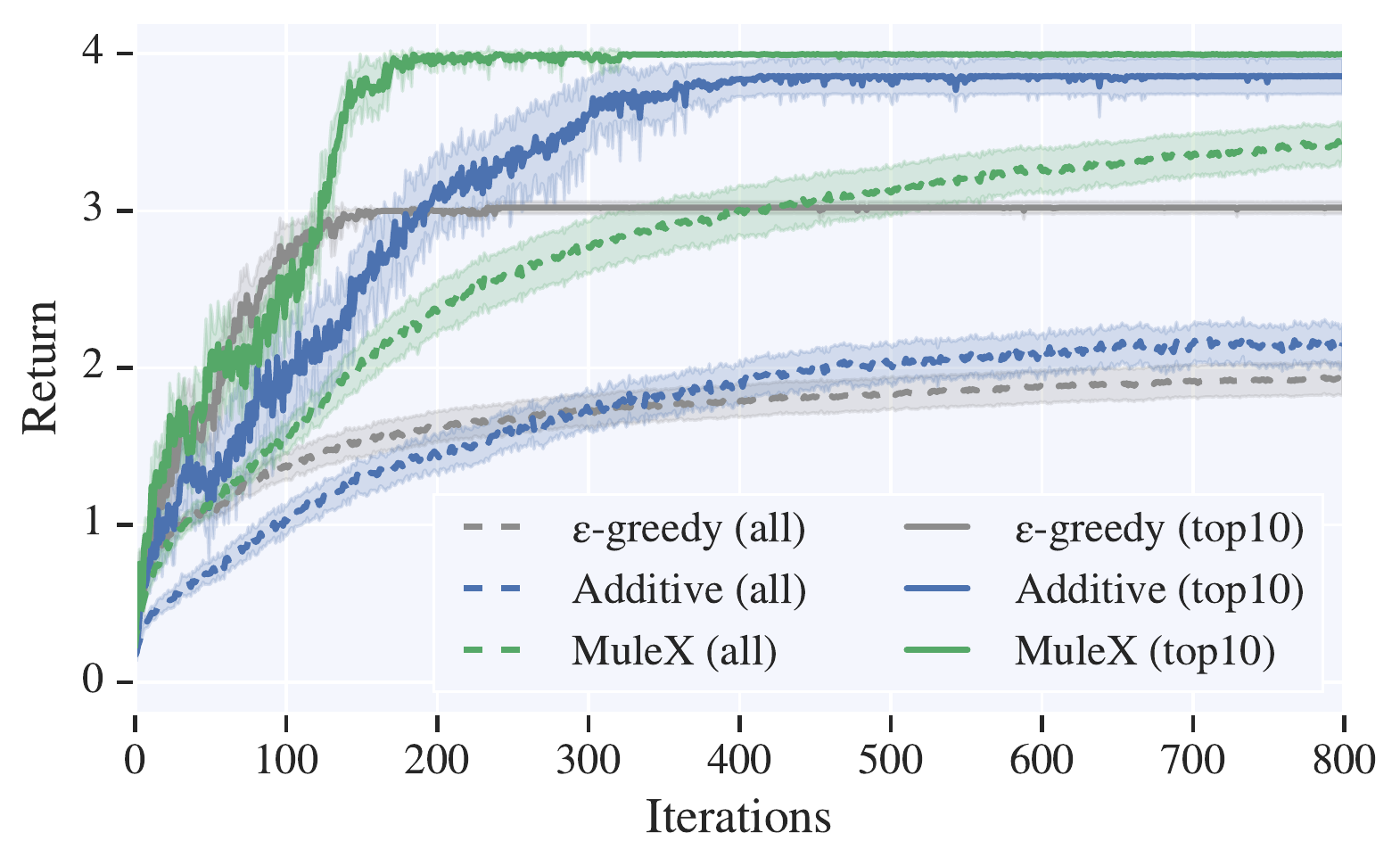}
  \end{center}
  \caption{Average performance of the agents throughout training. \textsc{MuleX} finds an optimal task-solving policy much faster.}\label{fig:xplor_curves}
\end{minipage}\hfill%
\begin{minipage}[t]{0.49\linewidth}
  \begin{center}
    \includegraphics[width=\linewidth]{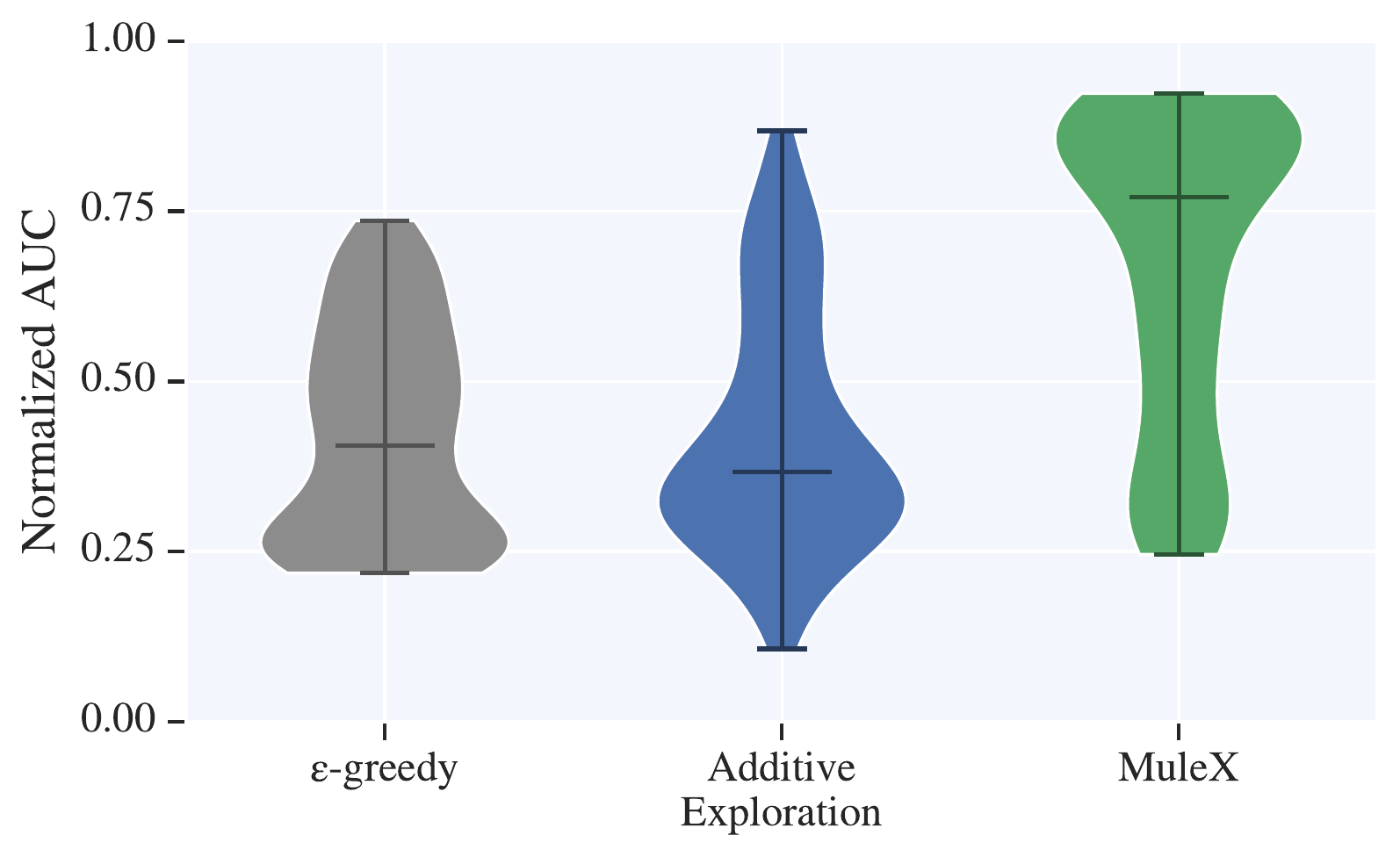}
  \end{center}
  \caption{Distribution of performance across hyperparameters. \textsc{MuleX} is significantly more robust to hyper-parameters.}\label{fig:xplor_violins}
\end{minipage}%
\end{figure}

\subsection{Faster optimal task-agent}
\label{sec:xp:faster}

First, we compare the return achieved by \textsc{MuleX}'s task-policy $\pi_\text{task}$ to that achieved by the {\it Additive} policy throughout training.
Figure~\ref{fig:xplor_curves} depicts these as curves showing the average over all $1000$ trials (dotted line) as well as the average of the best $10$ trials (solid line) according to the area under the curve (AUC).
As can be seen, \textsc{MuleX}'s task-policy reaches the highest return about twice as quickly as the {\it Additive} policy.
In the average case too, \textsc{MuleX}'s task-policy reaches higher return significantly quicker than the {\it Additive} policy, because it can focus on solving the task alone from the beginning.
This shows that learning separate policies for separate purposes is beneficial over learning a single policy with multiple, entangled purposes.

\subsection{Robustness to hyperparameters}
\label{sec:xp:robust}

Some RL methods can be very sensitive to hyperparameters, which implies that they need extra care to be properly tuned in order to perform at their best.

To investigate the robustness to hyperparameters we consider again the AUC.
For better interpretability we normalize it by the AUC of an ideal agent which would immediately obtain an ideal return from the first iteration on.
A normalized AUC of 1 represents this ideal agent.
We plot the density of normalized AUCs over the 1000 runs of the hyperparameter sweep on a violin plot in Figure~\ref{fig:xplor_violins}, including markers for the best, worst, and median performances.
These plots visualize the distribution of performances over all trials, and give a sense of robustness to hyperparameter values.
As we can see, even with such simple setup, the \textsc{MuleX} approach is significantly simpler to tune than the classical {\it Additive} one.

\subsection{Robustness of the task-policy w.r.t. initial states}
\label{sec:xp:s0}

\begin{wrapfigure}{r}{0.5\textwidth}
  \vspace{-20pt}
  \begin{center}
    \includegraphics[width=1.0\linewidth]{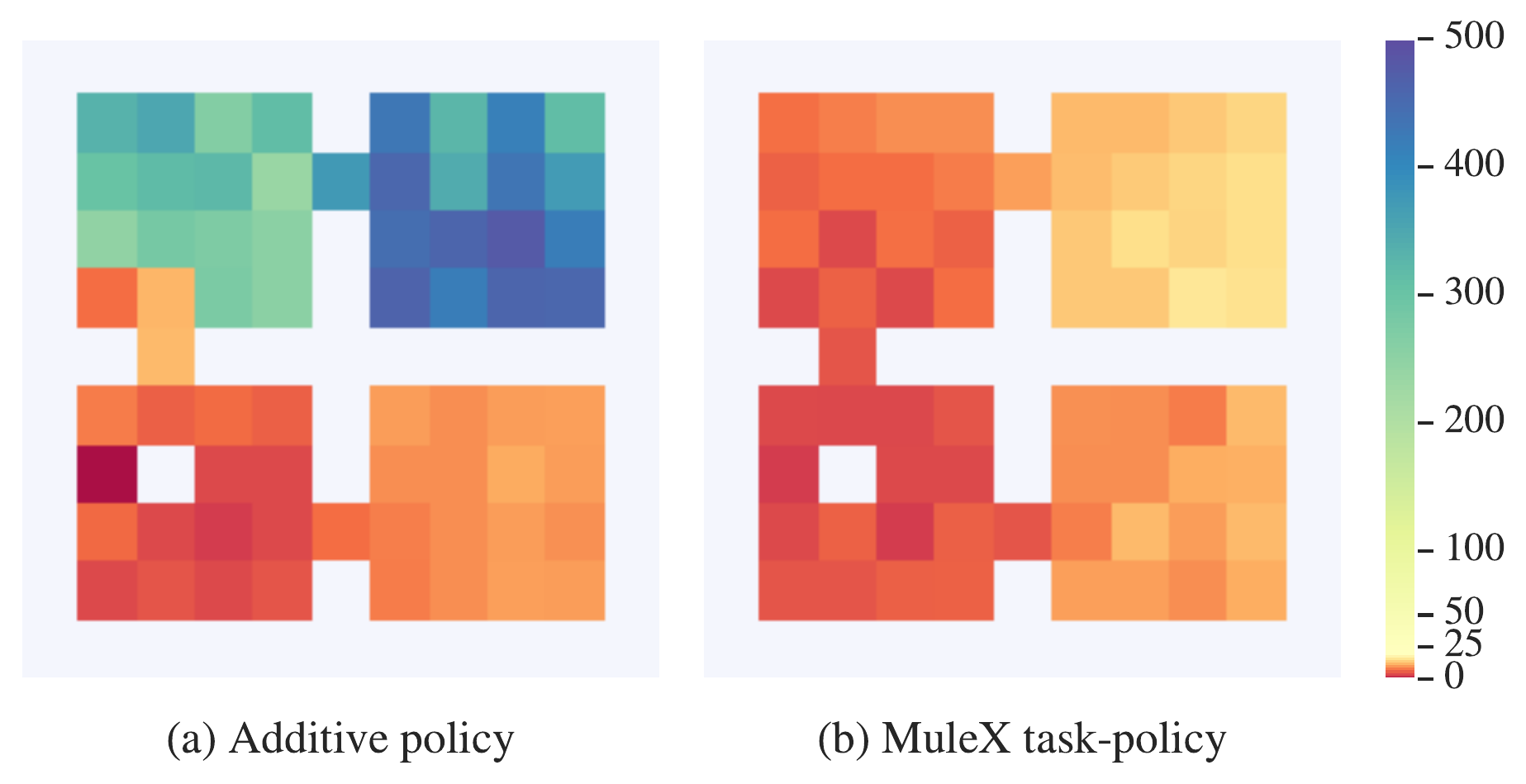}
  \end{center}
  \caption{Steps required to reach the goal from each starting location once all the keys and the extra bonus have been collected, averaged over 10 runs.}\label{fig:xplor_heatmaps}
  \vspace{-10pt}
\end{wrapfigure}

Another advantage of our framework is that the task-policy is inherently more robust for two main reasons:

\PAR{Continuous exploration:} By having distinct task and exploration policies, it is straightforward to keep exploring around well-trodden trajectories.
In contrast, \emph{Additive} explores less around the optimal trajectory over time.

\PAR{Immediate task solving:} When getting into rarely visited states,
the policy of the {\it Additive} approach exhibits explorative behaviour, because its reward is dominated by the high exploration reward that can be observed in those states.
On the other hand, the task-policy learned by \textsc{MuleX} is not contaminated by the exploration bonus and thus still learns to solve the task, \textit{i.e.}\ to ``get back to the optimal trajectory.''

We experimentally demonstrate this intuition by starting the best final trained policies in all possible states where both doors are open and the extra bonus is collected.
In this situation, the top two rooms are far off the optimal trajectory.
We then count how many steps are required for the policy to reach the goal starting from there.
The result is shown in Figure~\ref{fig:xplor_heatmaps} as a heatmap over all these starting states.
Note that the \textit{Additive}-exploration agent still wants to explore in the rarely seen top half of the map, whereas our \textsc{MuleX} agent's task policy goes straight to the goal, no matter how far off the optimal trajectory it starts.

\subsection{The necessity for all policies to act}
\label{subsec:all_policies}
\begin{wrapfigure}{r}{0.5\textwidth}
  \vspace{-20pt}
  \begin{center}
    \includegraphics[width=1.0\linewidth]{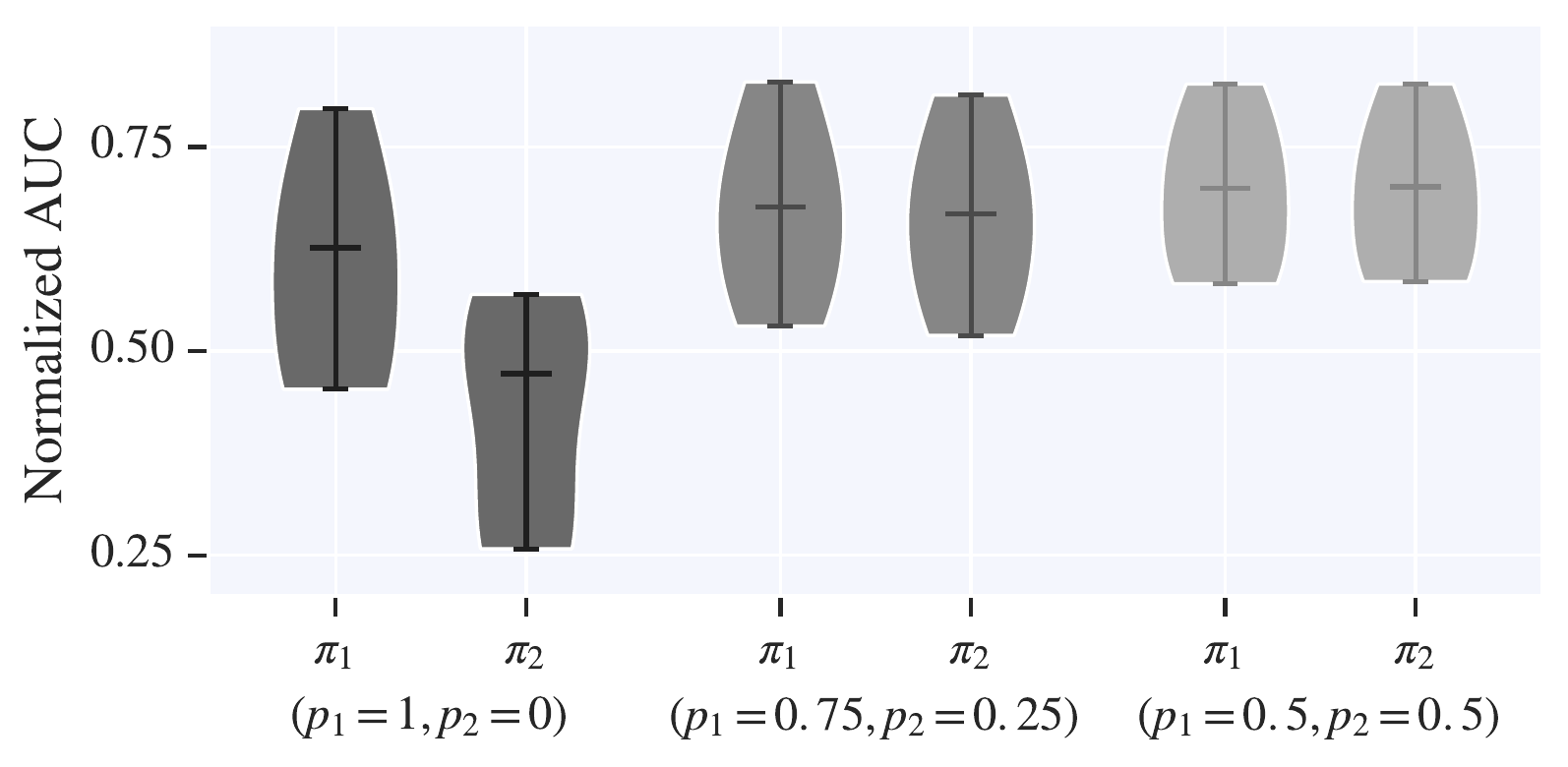}
  \end{center}
  \caption{Performance of acting probabilities.}\label{fig:offpolicy}
\end{wrapfigure}

Deciding which policy should act in the environment in order to collect transitions is an integral part of \textsc{MuleX}.
One could imagine that, in simple environments or when the policies are roughly aligned, it could be enough for the exploration policy to act, and the task policy $\pi_\text{task}$ could be learned completely offline.
It is well-known that offline Q-learning with function approximators can lead to overestimation of $Q_{\text{task}}(s,a)$ for some state action pairs and thus seriously impact the performance of the task policy \citep{fujimoto2019where}.
We experimentally demonstrate this by instantiating \textsc{MuleX} with two policies $\pi_1$ and $\pi_2$ which both optimize for the task-reward only using $\epsilon$-greedy exploration.
Both policies are trained from the same data and using the same reward.
We consider various start probabilities $p_1$ and $p_2$ for the acting strategy.
Figure~\ref{fig:offpolicy} confirms that even in a simple environment like Montezuminha, each policy must act to correct possible overestimation of its $Q$ function.

\subsection{Varying environment factors}

We now explicitly control each of the factors of the environment mentioned at the beginning of Section~\ref{sec:xp} which affect the exploration properties.

\subsubsection{Increased reward sparsity (room size)}

\begin{figure}[t]
  \begin{center}
    \includegraphics[width=1.0\linewidth]{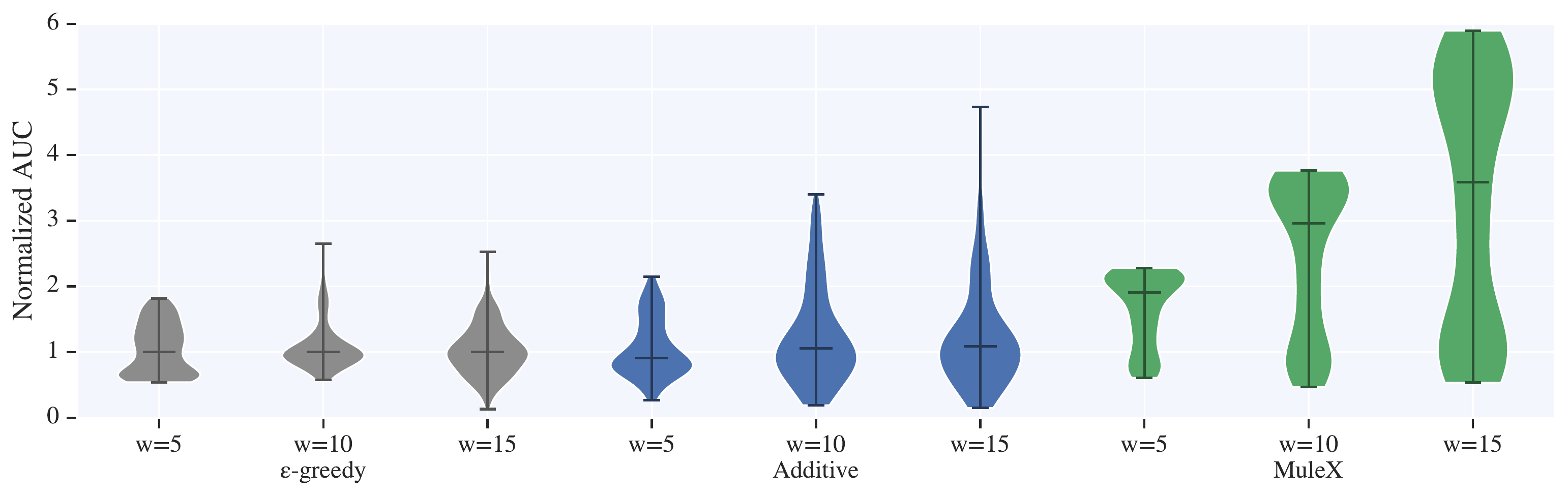}
  \end{center}
  \caption{Varying the task-reward sparsity by increasing room size $w$. For a given $w$, the AUC is normalized by the median AUC of the {\it $\epsilon$-greedy} baseline.}\label{fig:xplor_w}
\end{figure}

In order to investigate the effect of (task-)reward sparsity, we grow the room size $w$ from the previously used $w=5$ to $10$ and $15$.
This has the effect that the space to be explored grows quadratically, while the length of the optimal trajectory grows linearly.
We thus extend the maximum episode length to $1000$ and $1500$ steps as well as the steps per iteration to $5000$ and $7500$ for all agents.

Figure~\ref{fig:xplor_w} shows the results of this experiment.
For each room size $w$, we normalize the scores such that the median AUC of {\it $\epsilon$-greedy} is one.
This means that the plot shows how much {\it Additive} and \textsc{MuleX} improve over {\it $\epsilon$-greedy}.
The best agents performing increasingly better than the {\it $\epsilon$-greedy} baseline confirms that using an exploration bonus becomes increasingly important as the rewards get sparser.
However, \textsc{MuleX} offers significant advantages over the {\it Additive} method.
First, its median runs get better as the room size increases, demonstrating the robustness of our approach.
Second, \textsc{MuleX} makes better use of the exploration bonus than {\it Additive}: while for $w=5$, the best agents of both methods perform similarly, the gap increases significantly in favor of \textsc{MuleX} for $w=15$.

\subsubsection{Misleading exploration (teleporting walls)}

\begin{figure}[t]
\begin{minipage}[t]{0.49\linewidth}
  \begin{center}
    \includegraphics[width=\linewidth]{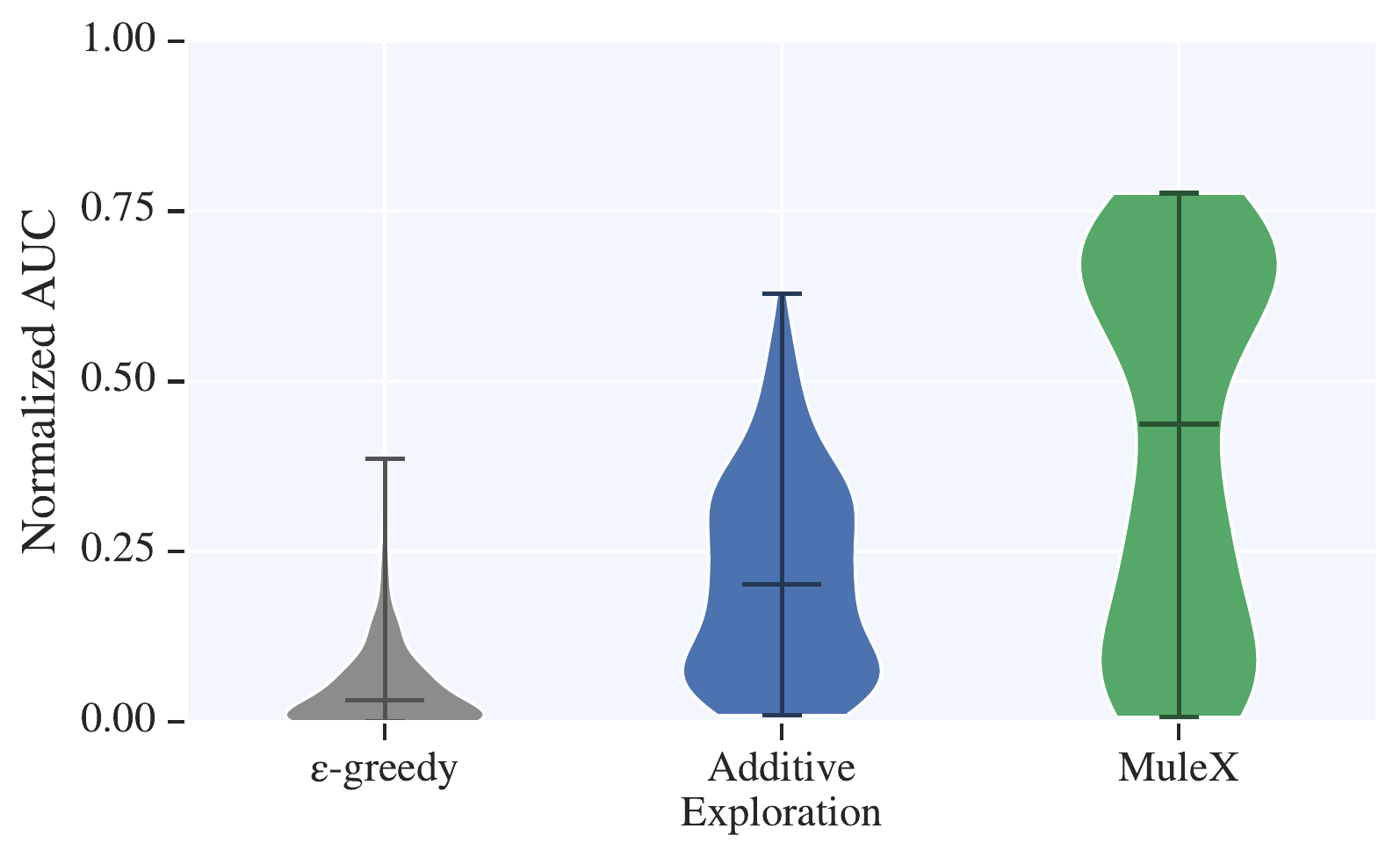}
  \end{center}
  \caption{Distribution of performance on the hard variant of Montezuminha.} \label{fig:hell_env_violins}
\end{minipage}\hfill%
\begin{minipage}[t]{0.49\linewidth}
  \begin{center}
    \includegraphics[width=1.0\linewidth]{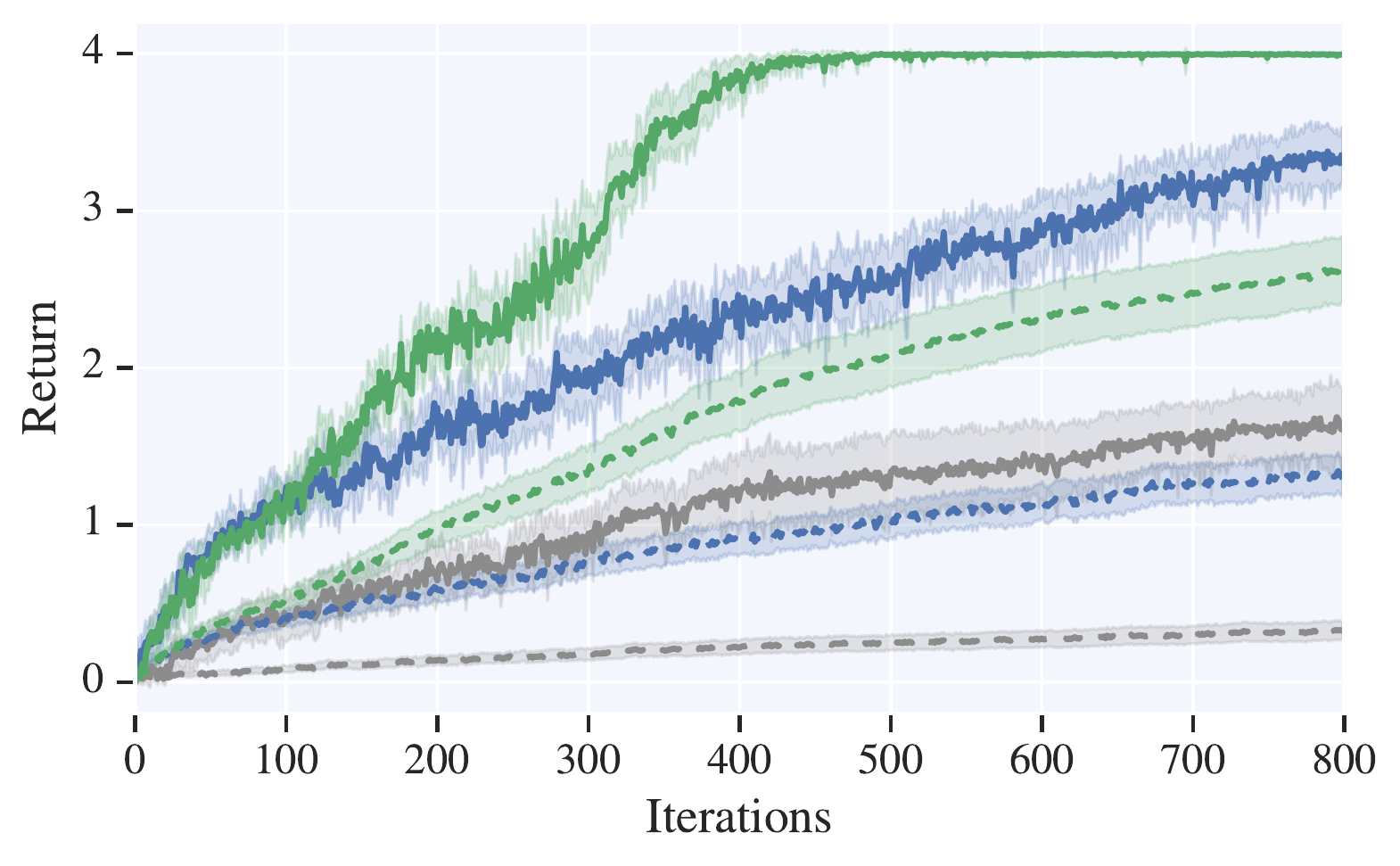}
  \end{center}
  \caption{Average performance throughout training on the hard variant of Montezuminha.} \label{fig:hell_env_curves}
\end{minipage}
\end{figure}

We also propose a significantly harder variant of Montezuminha, where the agent is teleported to a rewardless parallel world whenever it hits a wall.
A more detailed description of this environment is given in the Appendix.
The main challenge here is that exploring further is not necessarily aligned with solving the task.
The results in Figures~\ref{fig:hell_env_violins}~and~\ref{fig:hell_env_curves} again demonstrate that \textsc{MuleX} significantly outperforms the other two baselines.

\subsubsection{Stochasticity (random ghost)}

\begin{figure}[t]
\begin{minipage}[t]{0.49\linewidth}
  \begin{center}
    \includegraphics[width=1.0\linewidth]{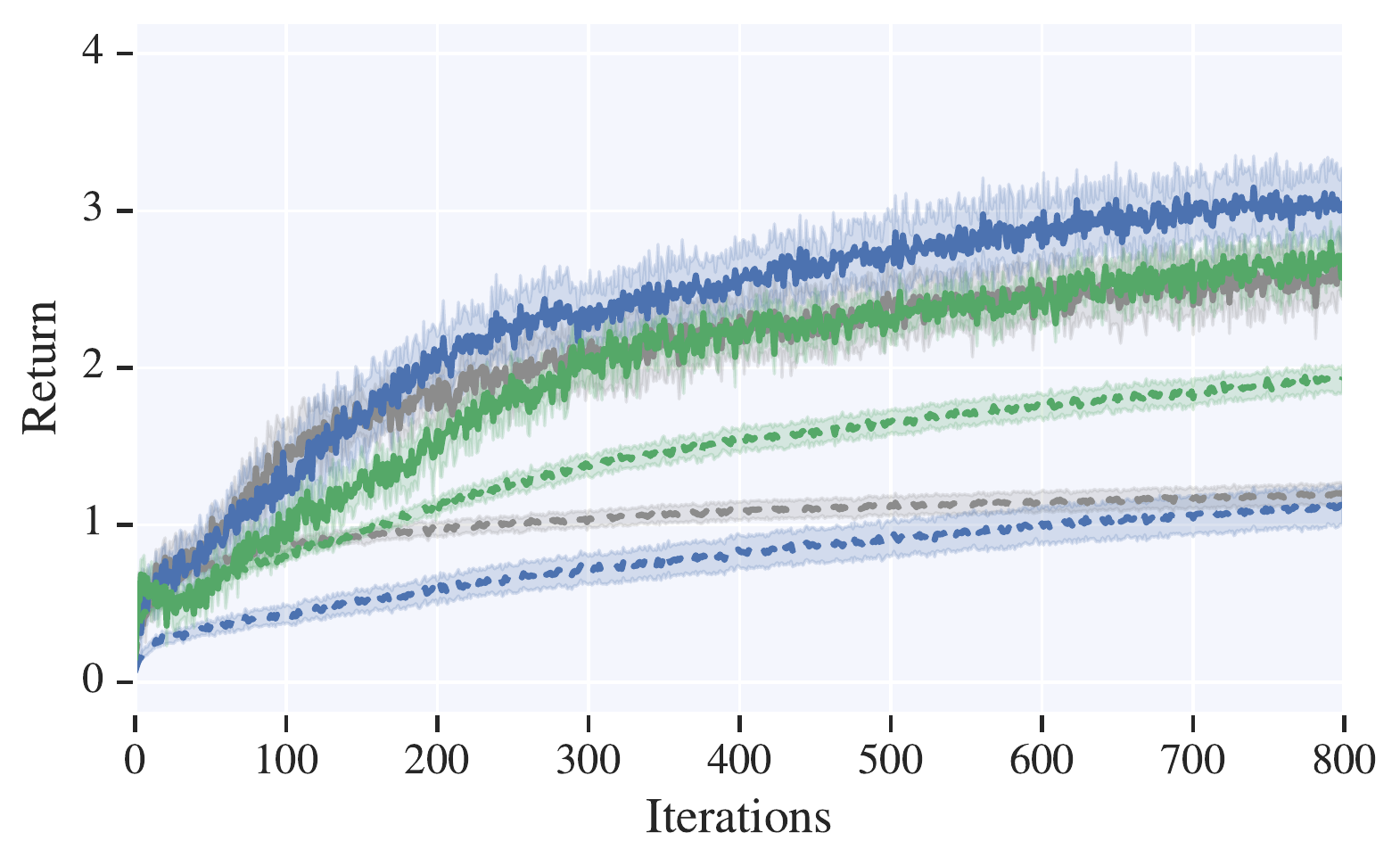}
  \end{center}
  \caption{Average performance of the agents through training in the stochastic environment.}\label{fig:xplor_sghosts}
\end{minipage}\hfill%
\begin{minipage}[t]{0.49\linewidth}
  \begin{center}
    \includegraphics[width=\linewidth]{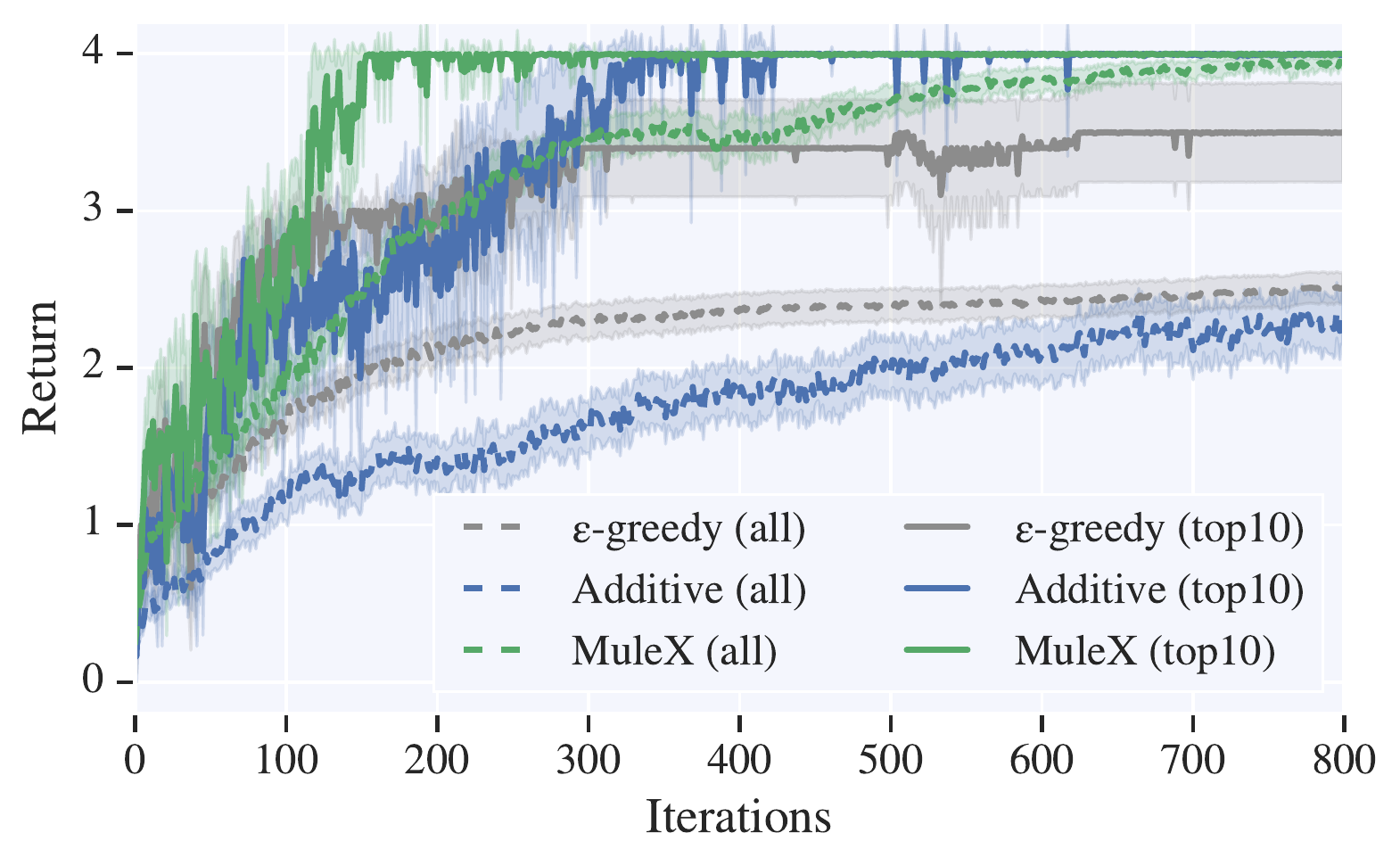}
  \end{center}
  \caption{Average performance of the agents in the textured environment and with pseudocounts.}\label{fig:xplor_texture}
\end{minipage}%
\end{figure}

While a deterministic environment can be useful for analyzing a method, RL aims at solving a wider range of problems, including stochastic environments.
This can pose a problem for exploration methods \citep{burda2018curiosity} as well as for algorithms \citep{ecoffet2019goexplore}.
For the sake of genericity, we make Montezuminha stochastic by introducing \emph{deadly ghosts} which move randomly and terminate the episode on contact, without giving negative reward.
We test the behaviour of \textsc{MuleX} and our baselines in this stochastic version of Montezuminha.
See the Appendix
for more details and the full results.
While the results in Figure~\ref{fig:xplor_sghosts} show that \textsc{MuleX} performs better than both baselines, it is evident that all three methods struggle, suggesting further work is needed in performing exploration in stochastic environments.

\subsubsection{Approximate exploration bonus (textures)}

Throughout this paper, we have used an oracle for the exploration bonus, because our goal is not to investigate exploration per-se, but rather to investigate new ways of integrating such bonus rewards into the training.
However, in most application scenarios, one does not have access to perfect (oracle) exploration boni, and it thus makes sense to evaluate how \textsc{MuleX} and \textit{Additive} agents behave under imperfect exploration boni.
For this, we use a textured version of Montezuminha and implement the SimHash-based exploration bonus proposed in \citet{tang2017exploration}.

The results are shown in Figure~\ref{fig:xplor_texture}, and we refer to the Appendix
for more details.
The difference of robustness with respect to hyperparameters is even more striking on this example:
\textsc{MuleX} can achieve the maximum return for almost any configuration while the {\it Additive} method requires some tuning of the hyper-parameters.
\textsc{MuleX} also solves the task much faster, as observed previously.

\section{Conclusion and future work}
\label{sec:conclusion}

\textsc{MuleX} is a new way to address the classic dilemma:
exploitation is disentangled from exploration by continuously optimizing a policy on the task reward, while performing exploration by acting according to a policy driven by a separate exploration objective. 
This new framework provides clear benefits both in terms of sample efficiency and robustness with respect to the initial state.

While we provide some insights on this new way of integrating bonus rewards, this is only a first step.
For example, we could consider elaborate actor selection strategies: intuitively, the task policy should act in well-explored part of the state space, whereas there is a need for more exploration in rarely visited states.
Another step could be applying these ideas to policy-gradient methods, which have been successfully used at scale.

Furthermore, \textsc{MuleX} seems like a natural candidate for life-long learning:
because it does not require any sort of annealing of exploration rewards,
it can constantly keep exploring without contaminating the task policy.

\clearpage

\bibliography{bib}

\begin{thebibliography}{26}
\providecommand{\natexlab}[1]{#1}
\providecommand{\url}[1]{\texttt{#1}}
\expandafter\ifx\csname urlstyle\endcsname\relax
  \providecommand{\doi}[1]{doi: #1}\else
  \providecommand{\doi}{doi: \begingroup \urlstyle{rm}\Url}\fi

\bibitem[Auer(2002)]{auer2002ucb}
Peter Auer.
\newblock {Using Confidence Bounds for Exploitation-Exploration Trade-offs}.
\newblock \emph{Journal of Machine Learning Research}, 3:\penalty0 397--422,
  2002.

\bibitem[Auer and Ortner(2007)]{NIPS2006_3052}
Peter Auer and Ronald Ortner.
\newblock Logarithmic online regret bounds for undiscounted reinforcement
  learning.
\newblock In B.~Sch\"{o}lkopf, J.~C. Platt, and T.~Hoffman, editors,
  \emph{Advances in Neural Information Processing Systems (NIPS)}, pages
  49--56. MIT Press, 2007.

\bibitem[Bellemare et~al.(2016)Bellemare, Srinivasan, Ostrovski, Schaul,
  Saxton, and Munos]{bellemare2016unifying}
Marc Bellemare, Sriram Srinivasan, Georg Ostrovski, Tom Schaul, David Saxton,
  and Remi Munos.
\newblock {Unifying Count-based Exploration and Intrinsic Motivation}.
\newblock In \emph{Advances in Neural Information Processing Systems (NIPS)},
  pages 1471--1479, 2016.

\bibitem[Bellemare et~al.(2017)Bellemare, Dabney, and
  Munos]{DBLP:journals/corr/BellemareDM17}
Marc~G Bellemare, Will Dabney, and R{\'e}mi Munos.
\newblock A distributional perspective on reinforcement learning.
\newblock In \emph{Proceedings of the International Conference on Machine
  Learning (ICML)}, pages 449--458. JMLR. org, 2017.

\bibitem[Bergstra and Bengio(2012)]{bergstra2012random}
James Bergstra and Yoshua Bengio.
\newblock {Random Search for Hyper-parameter Optimization}.
\newblock \emph{Journal of Machine Learning Research}, 13\penalty0
  (Feb):\penalty0 281--305, 2012.

\bibitem[Brafman and Tennenholtz(2002)]{brafman2002r}
Ronen~I Brafman and Moshe Tennenholtz.
\newblock {R-max - a General Polynomial Time Algorithm for Near-optimal
  Reinforcement Learning}.
\newblock \emph{Journal of Machine Learning Research}, 3\penalty0
  (Oct):\penalty0 213--231, 2002.

\bibitem[Bubeck and Cesa{-}Bianchi(2012)]{bubeck2012regret}
S{\'{e}}bastien Bubeck and Nicol{\`{o}} Cesa{-}Bianchi.
\newblock {Regret Analysis of Stochastic and Nonstochastic Multi-armed Bandit
  Problems}.
\newblock \emph{Foundations and Trends in Machine Learning}, 5\penalty0
  (1):\penalty0 1--122, 2012.

\bibitem[Burda et~al.(2019{\natexlab{a}})Burda, Edwards, Pathak, Storkey,
  Darrell, and Efros]{burda2018curiosity}
Yuri Burda, Harri Edwards, Deepak Pathak, Amos Storkey, Trevor Darrell, and
  Alexei~A. Efros.
\newblock Large-scale study of curiosity-driven learning.
\newblock In \emph{Proceedings of the International Conference on Learning
  Representations (ICLR)}, 2019{\natexlab{a}}.

\bibitem[Burda et~al.(2019{\natexlab{b}})Burda, Edwards, Storkey, and
  Klimov]{burda2018exploration}
Yuri Burda, Harrison Edwards, Amos Storkey, and Oleg Klimov.
\newblock {Exploration by random network distillation}.
\newblock In \emph{Proceedings of the International Conference on Learning
  Representations (ICLR)}, 2019{\natexlab{b}}.

\bibitem[Castro et~al.(2018)Castro, Moitra, Gelada, Kumar, and
  Bellemare]{castro18dopamine}
Pablo~Samuel Castro, Subhodeep Moitra, Carles Gelada, Saurabh Kumar, and
  Marc~G. Bellemare.
\newblock {Dopamine: A Research Framework for Deep Reinforcement Learning}.
\newblock \emph{CoRR}, abs/1812.06110, 2018.

\bibitem[Colas et~al.(2018)Colas, Sigaud, and Oudeyer]{colas2018gep}
C{\'e}dric Colas, Olivier Sigaud, and Pierre-Yves Oudeyer.
\newblock {GEP-PG: Decoupling Exploration and Exploitation in Deep
  Reinforcement Learning Algorithms}.
\newblock In \emph{Proceedings of the International Conference on Machine
  Learning (ICML)}, 2018.

\bibitem[Ecoffet et~al.(2019)Ecoffet, Huizinga, Lehman, Stanley, and
  Clune]{ecoffet2019goexplore}
Adrien Ecoffet, Joost Huizinga, Joel Lehman, Kenneth~O. Stanley, and Jeff
  Clune.
\newblock {Go-Explore: a New Approach for Hard-Exploration Problems}.
\newblock \emph{CoRR}, abs/1901.10995, 2019.

\bibitem[Fortunato et~al.(2018)Fortunato, Azar, Piot, Menick, Osband, Graves,
  Mnih, Munos, Hassabis, Pietquin, Blundell, and Legg]{fortunato2017noisy}
Meire Fortunato, Mohammad~Gheshlaghi Azar, Bilal Piot, Jacob Menick, Ian
  Osband, Alexander Graves, Vlad Mnih, Remi Munos, Demis Hassabis, Olivier
  Pietquin, Charles Blundell, and Shane Legg.
\newblock {Noisy Networks for Exploration}.
\newblock In \emph{Proceedings of the International Conference on
  Representation Learning (ICLR)}, 2018.

\bibitem[Fujimoto et~al.(2019)Fujimoto, Meger, and Precup]{fujimoto2019where}
Scott Fujimoto, David Meger, and Doina Precup.
\newblock Where off-policy deep reinforcement learning fails, 2019.
\newblock URL \url{https://openreview.net/forum?id=S1zlmnA5K7}.

\bibitem[Geist and Pietquin(2011)]{geist2011managing}
Matthieu Geist and Olivier Pietquin.
\newblock Managing uncertainty within the ktd framework.
\newblock In \emph{Active Learning and Experimental Design workshop in
  conjunction with AISTATS 2010}, pages 157--168, 2011.

\bibitem[Hessel et~al.(2018)Hessel, Modayil, Van~Hasselt, Schaul, Ostrovski,
  Dabney, Horgan, Piot, Azar, and Silver]{DBLP:journals/corr/abs-1710-02298}
Matteo Hessel, Joseph Modayil, Hado Van~Hasselt, Tom Schaul, Georg Ostrovski,
  Will Dabney, Dan Horgan, Bilal Piot, Mohammad Azar, and David Silver.
\newblock Rainbow: Combining improvements in deep reinforcement learning.
\newblock In \emph{Proceedings of the AAAI Conference on Artificial
  Intelligence (AAAI)}, 2018.

\bibitem[Jaderberg et~al.(2018)Jaderberg, Czarnecki, Dunning, Marris, Lever,
  Castaneda, Beattie, Rabinowitz, Morcos, Ruderman, et~al.]{jaderberg2018human}
Max Jaderberg, Wojciech~M Czarnecki, Iain Dunning, Luke Marris, Guy Lever,
  Antonio~Garcia Castaneda, Charles Beattie, Neil~C Rabinowitz, Ari~S Morcos,
  Avraham Ruderman, et~al.
\newblock {Human-level Performance in First-person Multiplayer Games with
  Population-based Deep Reinforcement Learning}.
\newblock \emph{arXiv preprint arXiv:1807.01281}, 2018.

\bibitem[Mnih et~al.(2015)Mnih, Kavukcuoglu, Silver, Rusu, Veness, Bellemare,
  Graves, Riedmiller, Fidjeland, Ostrovski, et~al.]{mnih2015human}
Volodymyr Mnih, Koray Kavukcuoglu, David Silver, Andrei~A Rusu, Joel Veness,
  Marc~G Bellemare, Alex Graves, Martin Riedmiller, Andreas~K Fidjeland, Georg
  Ostrovski, et~al.
\newblock {Human-level Control Through Deep Reinforcement Learning}.
\newblock \emph{Nature}, 518\penalty0 (7540):\penalty0 529, 2015.

\bibitem[Mnih et~al.(2016)Mnih, Badia, Mirza, Graves, Lillicrap, Harley,
  Silver, and Kavukcuoglu]{mnih2016asynchronous}
Volodymyr Mnih, Adria~Puigdomenech Badia, Mehdi Mirza, Alex Graves, Timothy
  Lillicrap, Tim Harley, David Silver, and Koray Kavukcuoglu.
\newblock {Asynchronous Methods for Deep Reinforcement Learning}.
\newblock In \emph{Proceedings of the International conference on machine
  learning (ICML)}, pages 1928--1937, 2016.

\bibitem[Plappert et~al.(2018)Plappert, Houthooft, Dhariwal, Sidor, Chen, Chen,
  Asfour, Abbeel, and Andrychowicz]{plappert2017parameter}
Matthias Plappert, Rein Houthooft, Prafulla Dhariwal, Szymon Sidor, Richard~Y.
  Chen, Xi~Chen, Tamim Asfour, Pieter Abbeel, and Marcin Andrychowicz.
\newblock {Parameter Space Noise for Exploration}.
\newblock In \emph{Proceedings of the International Conference on Learning
  Representations (ICLR)}, 2018.

\bibitem[Schaul et~al.(2016)Schaul, Quan, Antonoglou, and
  Silver]{DBLP:journals/corr/SchaulQAS15}
Tom Schaul, John Quan, Ioannis Antonoglou, and David Silver.
\newblock Prioritized experience replay.
\newblock In \emph{Proceedings of the International Conference on Learning
  Representations (ICLR)}, 2016.

\bibitem[Sehnke et~al.(2010)Sehnke, Osendorfer, R{\"u}ckstie{\ss}, Graves,
  Peters, and Schmidhuber]{sehnke2010parameter}
Frank Sehnke, Christian Osendorfer, Thomas R{\"u}ckstie{\ss}, Alex Graves, Jan
  Peters, and J{\"u}rgen Schmidhuber.
\newblock {Parameter-exploring Policy Gradients}.
\newblock \emph{Neural Networks}, 23\penalty0 (4):\penalty0 551--559, 2010.

\bibitem[Strehl and Littman(2008)]{Strehl08}
Alexander~L. Strehl and Michael~L. Littman.
\newblock An analysis of model-based interval estimation for markov decision
  processes.
\newblock \emph{J. Comput. Syst. Sci.}, 74\penalty0 (8):\penalty0 1309--1331,
  2008.

\bibitem[Sutton and Barto(2018)]{sutton2018reinforcement}
Richard~S Sutton and Andrew~G Barto.
\newblock \emph{{Reinforcement Learning: An Introduction}}.
\newblock MIT press, 2018.

\bibitem[Tang et~al.(2017)Tang, Houthooft, Foote, Stooke, Chen, Duan, Schulman,
  DeTurck, and Abbeel]{tang2017exploration}
Haoran Tang, Rein Houthooft, Davis Foote, Adam Stooke, OpenAI~Xi Chen, Yan
  Duan, John Schulman, Filip DeTurck, and Pieter Abbeel.
\newblock {\# Exploration: A Study of Count-based Exploration for Deep
  Reinforcement Learning}.
\newblock In \emph{Advances in Neural Information Processing Systems (NIPS)},
  pages 2753--2762, 2017.

\bibitem[Vemula et~al.(2019)Vemula, Sun, and Bagnell]{vemula2019contrasting}
Anirudh Vemula, Wen Sun, and J~Andrew Bagnell.
\newblock {Contrasting Exploration in Parameter and Action Space: A
  Zeroth-Order Optimization Perspective}.
\newblock In \emph{Proceedings of the International Conference on Artificial
  Intelligence and Statistics (AISTATS)}, 2019.

\end{thebibliography}
{\small \bibliographystyle{plainnat}}

\clearpage

\appendix

\section{Additional experiments}

\subsection{Rainbow}
\label{sec:app:rainbow}

\begin{figure}[t]
\begin{minipage}[t]{0.49\linewidth}
  \begin{center}
    \includegraphics[width=\linewidth]{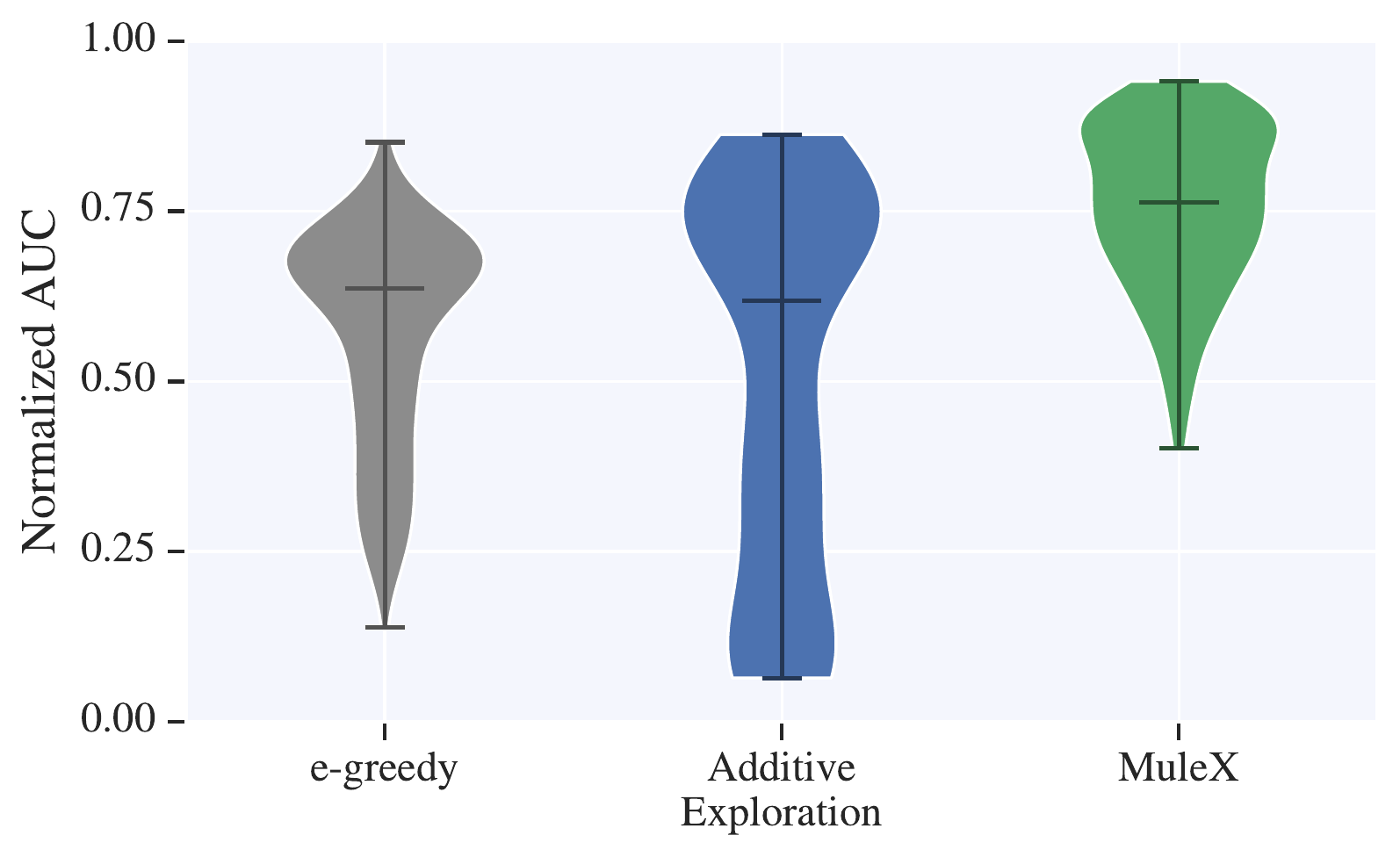}
  \end{center}
  \caption{Distribution of performance across hyperparameters using Rainbow inplace of DQN.} \label{fig:rainbow_violins}
\end{minipage}\hfill%
\begin{minipage}[t]{0.49\linewidth}
  \begin{center}
    \includegraphics[width=1.0\linewidth]{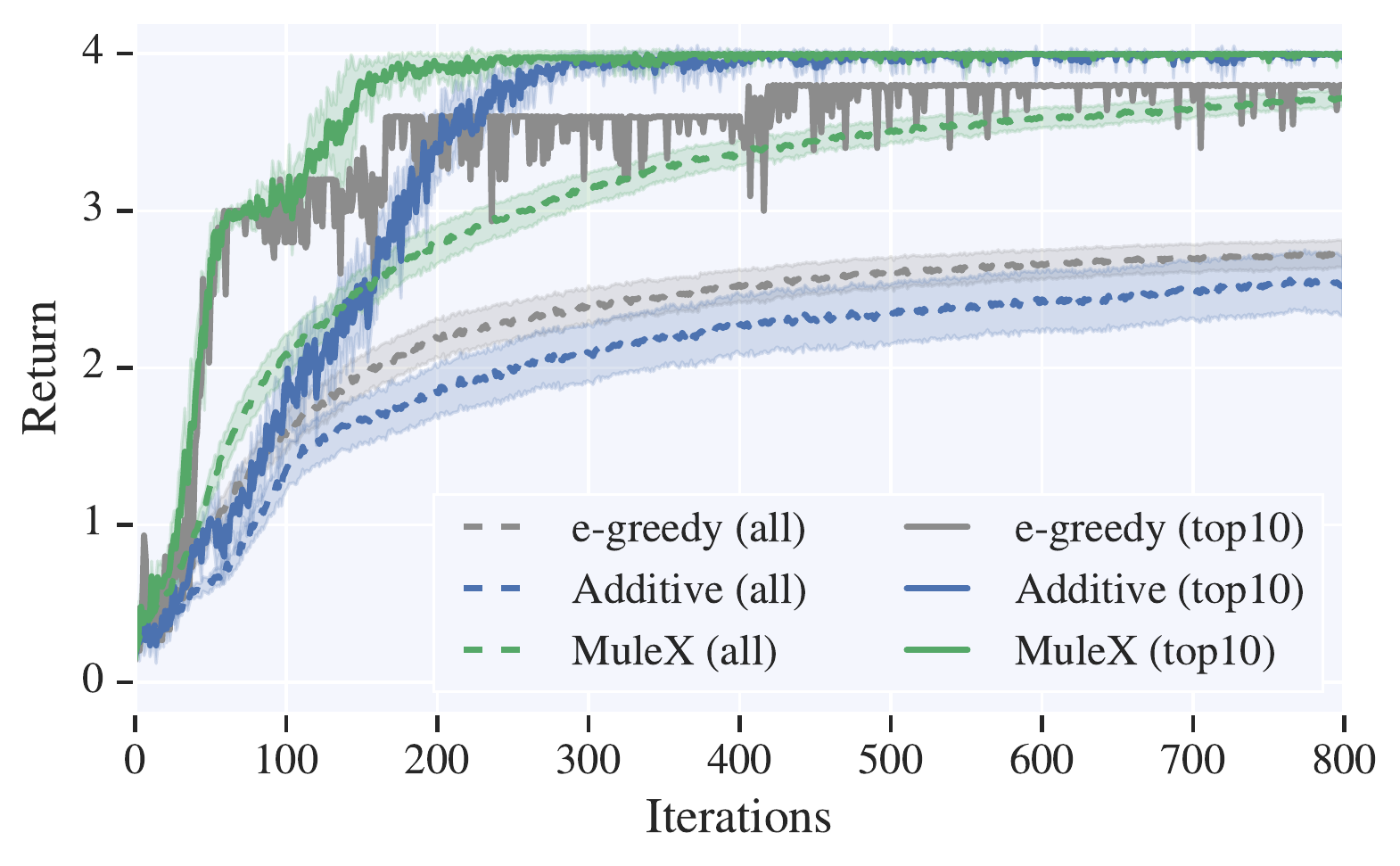}
  \end{center}
  \caption{Average performance of the Rainbow agents throughout training.} \label{fig:rainbow_curves}
\end{minipage}
\end{figure}

Rainbow \citep{DBLP:journals/corr/abs-1710-02298} is a combination of extensions that improve over the original DQN \citep{mnih2015human}. We use the Dopamine implementation of Rainbow which includes the following extensions: n-step returns, prioritized experience replay   \citep{DBLP:journals/corr/SchaulQAS15} and C51 distributional RL  \citep{DBLP:journals/corr/BellemareDM17}.

Prioritized experience replay was shown to be effective for MDPs with sparse and stationary rewards. This remark drives our choice for \textsc{MuleX} where the task policy is learned using Rainbow while the exploration policy, which is learned from dense and non-stationary rewards is trained using standard DQN. Note that the {\it Additive} exploration method in conjunction with Rainbow is an important baseline although the additive reward does not meet the stationarity assumption.

We directly compare the three methods, namely {\it $\epsilon$-greedy}, {\it additive exploration}, \textsc{MuleX}, using Rainbow with the results obtained with standard DQN, given in Figures~\ref{fig:xplor_curves},\ref{fig:xplor_violins}.
Figures~\ref{fig:rainbow_curves},\ref{fig:rainbow_violins} show that all methods benefit from the extensions provided in Rainbow.
Together with faster learning curves, the most notable impact on \textsc{MuleX} is an additional strong increase of robustness to hyper-parameters as the minimum AUC increased significantly.

\subsection{Hard variant of the environment}
\label{sec:hard_variant}

\begin{figure}[b]
    \begin{center}
    \includegraphics[width=0.6\linewidth]{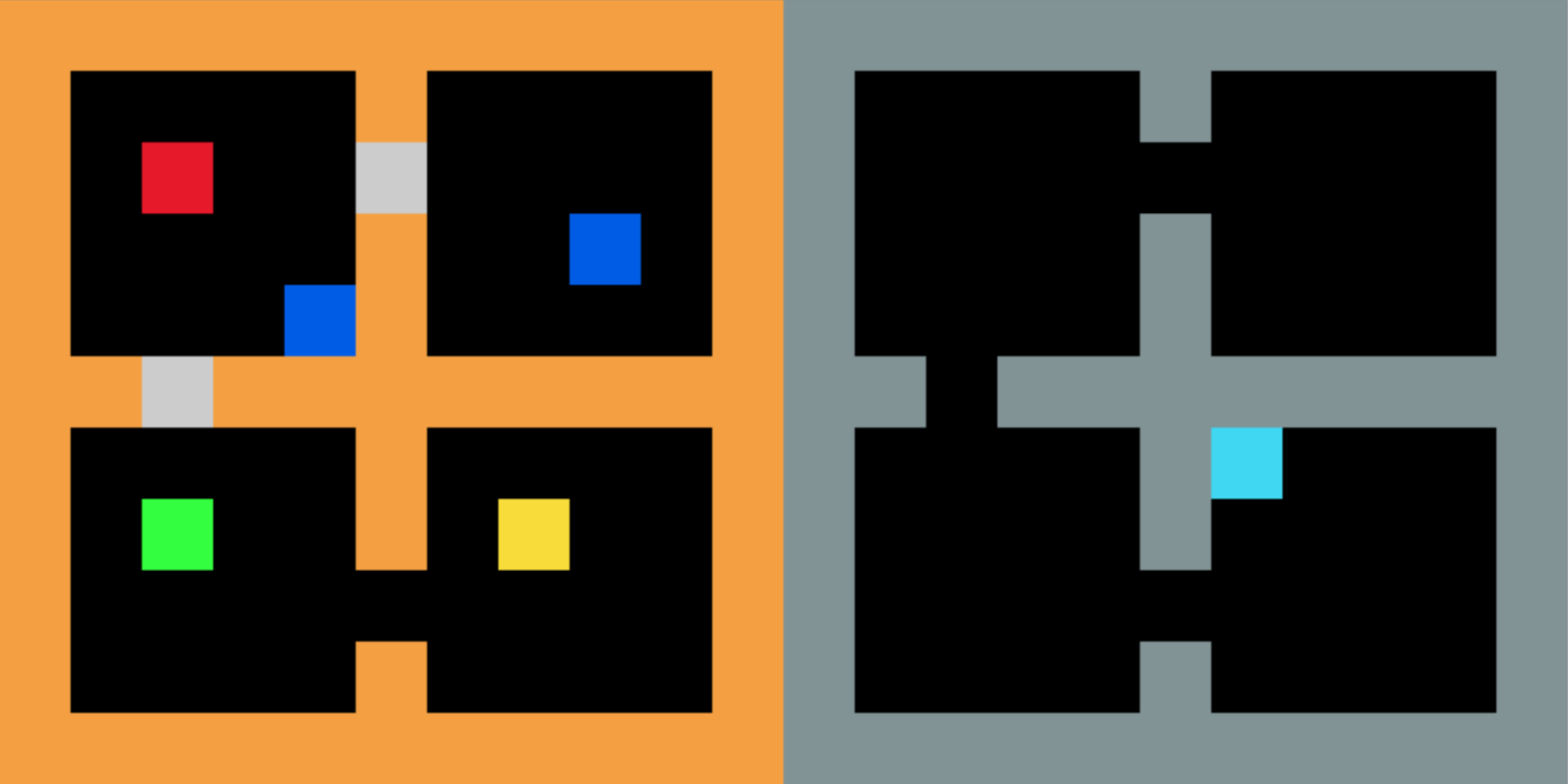}
    \end{center}
    \caption{The full view of the hard variant of Montezuminha. The set of 4 left rooms correspond to the original Montezuminha environment as described in Figure~\ref{fig:env}, except the golden walls now teleport the agent to the corresponding location on the right side. Reaching the light blue location on the right side is then the only way for the agent to get back to the left side and solve the task.}
    \label{fig:hard_env}
\end{figure}

In the variants of Montezuminha considered so far, exploration is always \emph{safe} in the sense that the exploration is very much aligned with solving the task, and in the worst case only wastes a bit of time.
We introduce a hard variant of Montezuminha, shown on Figure~\ref{fig:hard_env}. When the agent is located in one of the rooms on the left side and touches the wall, it gets teleported to the same location but on the right side. Then, the only way to escape is to reach the object located in the bottom right room which teleports the agent back to the initial state on the left.
This parallel world on the right is only distracting when it comes to solving the task but makes the exploration task much harder.

We show on Figures~\ref{fig:hell_env_violins} and \ref{fig:hell_env_curves} the performance of \textsc{MuleX} compared to the baselines. 
In this variant of Montezuminha, it is now extremely difficult to discover new rewards just by chance which explains the poor performance of the agent learned using $\epsilon$-greedy exploration. Compared to the additive baseline, \textsc{MuleX} trains faster and still manages to reach the best return possible within the budget of 800 training iterations.

\subsection{Details and full results on stochastic environment}
\label{sec:app:ghosts}

A ghost has a current moving direction, which has a probability of 25\% to randomly change at every step.
This means that it is possible to reason at least a little about these random ghosts.
The ghost can also walk through doors into other rooms, if the doors are open.

We make the environment stochastic by putting one single ghost into the same room the player starts in.
We also increase the room size to $w=10$ as otherwise the chance of collision in the first room is much too high.

\begin{figure}[t]
\begin{minipage}[t]{0.49\linewidth}
  \begin{center}
    \includegraphics[width=\linewidth]{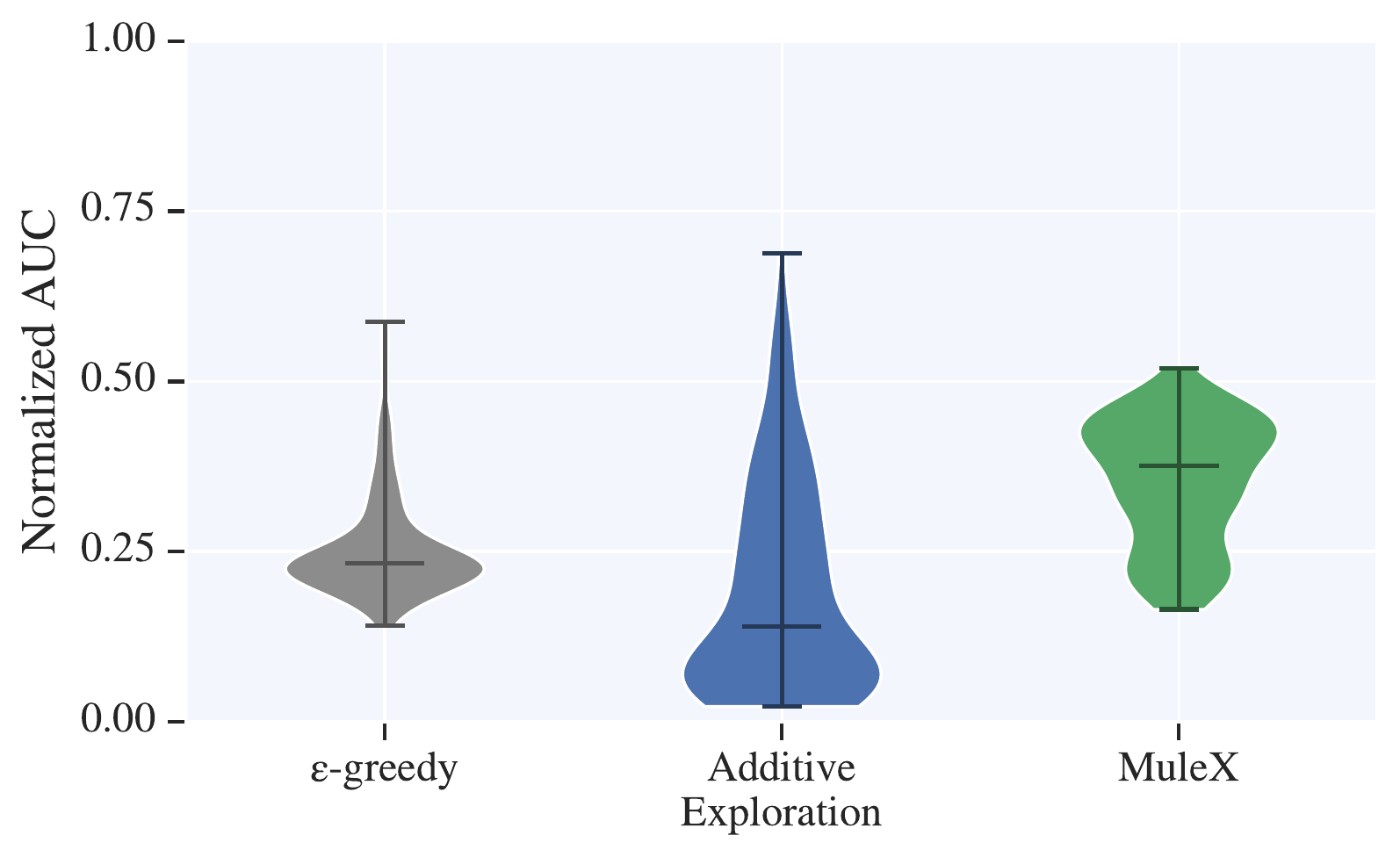}
  \end{center}
  \caption{Distribution of performance with a ghost.} \label{fig:sghosts_violins}
\end{minipage}\hfill%
\begin{minipage}[t]{0.49\linewidth}
  \begin{center}
    \includegraphics[width=1.0\linewidth]{images/xplor_sghosts_curves.pdf}
  \end{center}
  \caption{Average performance throughout training with a ghost.} \label{fig:sghosts_curves}
\end{minipage}
\end{figure}

See Figure~\ref{fig:sghosts_violins} and Figure~\ref{fig:sghosts_curves} for all results.

\subsection{Full results on textured version with pseudocounts}
\label{sec:app:textures}

\begin{figure}[b]
  \begin{center}
    \includegraphics[width=0.5\linewidth]{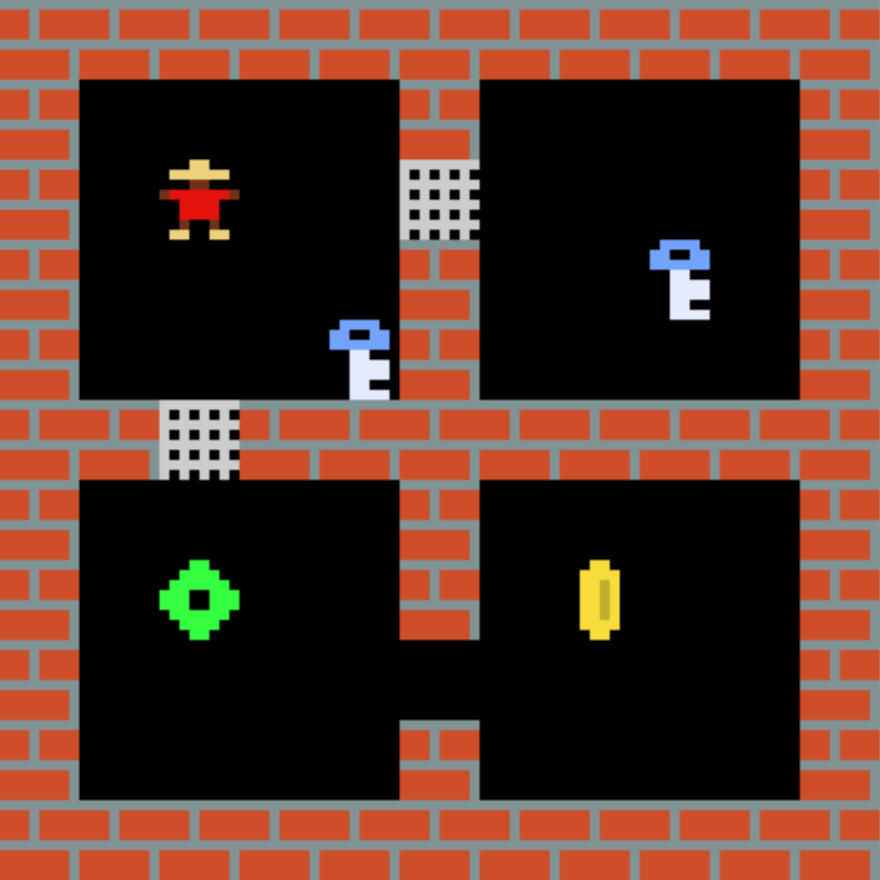}
  \end{center}
  \caption{The textured version of the Montezuminha environment.}\label{fig:montezuminha_texture}
\end{figure}

Each type of cell has a texture of $8\times8$ pixels associated to it, and the status bar (showing collected items) shows the same texture as used in the room view.
The textured environment is shown in Figure~\ref{fig:montezuminha_texture}
Because of the increase in size, we also increase the convolutional body's capacity of the network slightly by adding a convolution and increasing filter sizes.

For the SimHash exploration bonus, we resize the input to $13\times13$, use 10 value bins, and project the result to a random code of size 256.
These settings ensure that it does not degenerate to an oracle reward, but could contain mistakes.

The textured environment is computationally much more demanding, and thus we use only a single random seed for each random hyperparameter, resulting in only 200 runs of each method (instead of 5 repeats resulting in 1000 runs for all other experiments).

\begin{figure}[t]
\begin{minipage}[t]{0.49\linewidth}
  \begin{center}
    \includegraphics[width=1.0\linewidth]{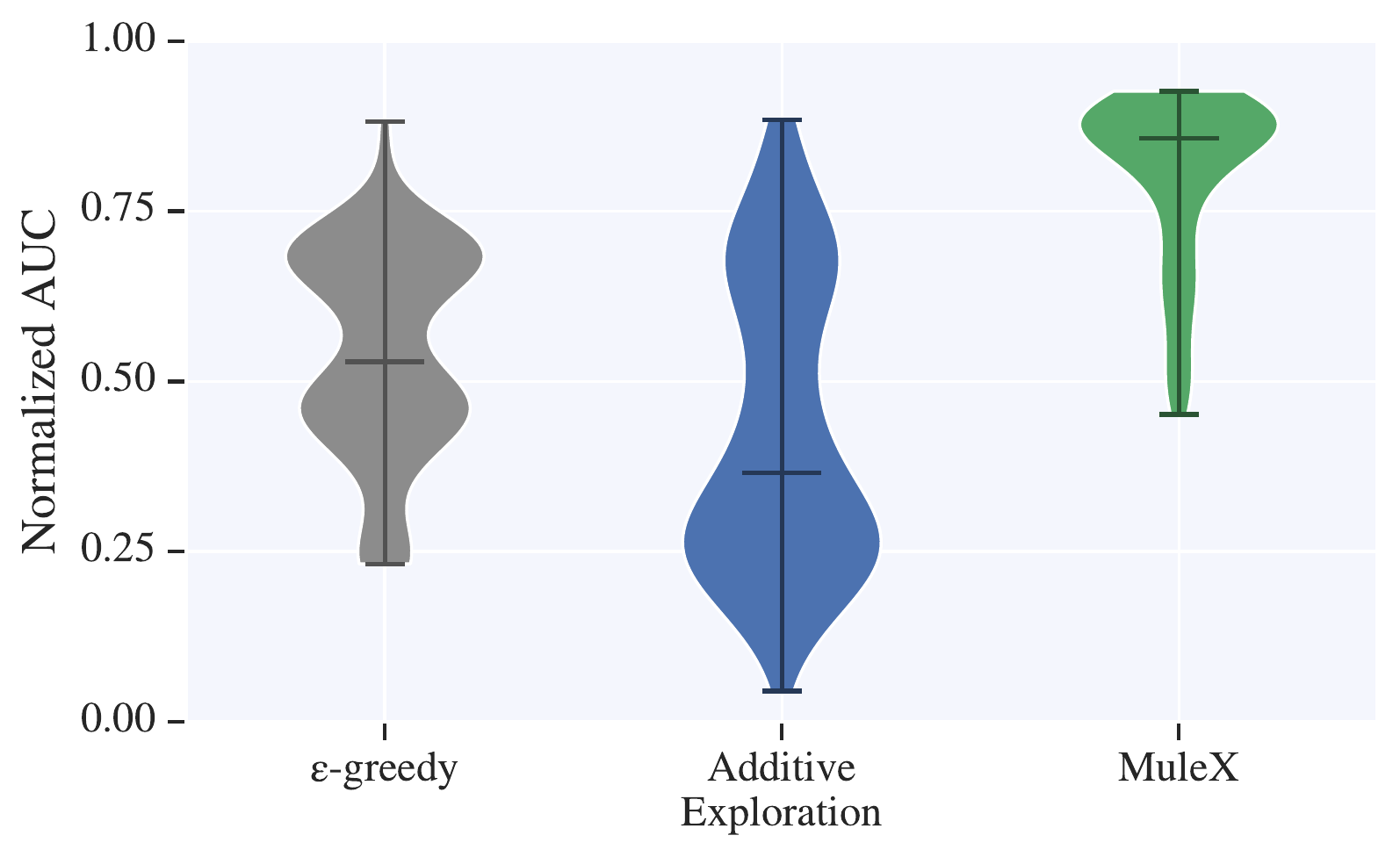}
  \end{center}
  \caption{Distribution of performance on the textured version and using pseudocounts.}\label{fig:texture_violins}
\end{minipage}\hfill%
\begin{minipage}[t]{0.49\linewidth}
  \begin{center}
    \includegraphics[width=\linewidth]{images/xplor_texture_curves.pdf}
  \end{center}
  \caption{Average performance of the agents through training on the textured version of Montezuminha and using pseudocounts.}\label{fig:texture_curves}
\end{minipage}%
\end{figure}

See Figure~\ref{fig:texture_violins} and Figure~\ref{fig:texture_curves}.

\end{document}